# Data-driven Intelligent Computational Design for Products: Method, Techniques, and Applications

Maolin Yang [a†], Pingyu Jiang [a†*], Tianshuo Zang [a], and Yuhao Liu [a]

*[a] State Key Laboratory for Manufacturing Systems Engineering, Xi'an Jiaotong University, Xi'an, Shaanxi 710054, China*

*† These authors contributed equally to this work*

*\* Correspond to: pjiang@mail.xjtu.edu.cn*

**Abstract**

Data-driven intelligent computational design (DICD) is a research hotspot emerged under the context of fast-developing artificial intelligence. It emphasizes on utilizing deep learning algorithms to extract and represent the design features hidden in historical or fabricated design process data, and then learn the combination and mapping patterns of these design features for the purposes of design solution retrieval, generation, optimization, evaluation, etc. Due to its capability of automatically and efficiently generating design solutions and thus supporting human-in-the-loop intelligent and innovative design activities, DICD has drawn the attentions from both academic and industrial fields. However, as an emerging research subject, there are still many unexplored issues that limit the development and application of DICD, such as specific dataset building, engineering design related feature engineering, systematic methods and techniques for DICD implementation in the entire product design process, etc. In this regard, a systematic and operable road map for DICD implementation from full-process perspective is established, including a general workflow for DICD project planning, an overall framework for DICD project implementation, the computing mechanisms for DICD implementation, key enabling technologies for detailed DICD implementation, and three application scenarios of DICD. The road map reveals the common mechanisms and calculation principles of existing DICD researches, and thus it can provide systematic guidance for the possible DICD applications that have not been explored.

**List of symbols**

| | |
|---|---|
| $g_c$ | The *c* th constraint contained in an intelligent design model. |
| *s* | A piece of sequential datum that represents a part of a design solution. |
| $w_i$ | The weight of $x_i$. |
| $x_i$ | The $i$ th design feature in $X$. |
| $y_j$ | The $j$ th design feature in $Y$. |
| *z* | A piece of noise. |
| *C* | The number of constraints contained in the intelligent design model. |
| *F* | The mapping function from $X$ to $Y$. |
| *G* | Gaussian distribution. |
| *M* | The number of design features in the output parts of all the training data for the intelligent design model. |
| *N* | The number of design features in the input parts of all the training data for the intelligent design model. |
| *P* | The number of dimensions of the feature space of $X$ for a supervised learning based retrieval model. |
| *Q* | The number of dimensions of the feature space of $Y$ for a supervised learning based retrieval model. |
| *R* | The number of dimensions of the feature space of both $X$ and $Y$ for a unsupervised learning based retrieval model. |
| $T_X$ | The preprocessing and feature engineering process that transforms $X_{Raw}$ to $X$. |
| $T_Y$ | The preprocessing and feature engineering process that transforms $Y_{Raw}$ to $Y$. |
| *W* | The set of the weights for all the design features in $X$. |
| *X* | The input part of a piece of training datum for the intelligent design model. |
| $X_{Raw}$ | The input part of a piece of raw design solution datum for the intelligent design model. |
| *Y* | The output part of a piece of training datum for the intelligent design model. |
| $Y_{Raw}$ | The output part of a piece of raw design solution datum for the intelligent design model. |
| $\varepsilon *$ | Stress. |
| $\sigma *$ | Strain. |
| $\Omega^X$ | The set of the design features in the input parts of all the training data for the intelligent design model. |
| $\Omega^Y$ | The set of the design features in the output parts of all the training data for the intelligent design model. |

**1. Introduction**

Product design can be roughly separated into Classical design, Design thinking, and Computational design according to the methods and techniques applied during design process (Maeda, 2019). Classical design reflects the mapping relations of Requirements → Functions → Behaviors → Structures, and it can be traced back to thousands of years ago when humans are making bows and arrows. Design thinking, e.g. TRIZ innovative design methods (Donnici *et al.*, 2022), is more of the patterns and ways of thinking for efficient design innovation, concluded from accumulated design experiences. Computational design emphasizes on utilizing mathematics and calculations to acquire the optimal design solutions, and it can be further separated into experience-driven (Costa *et al.*, 2012), model-driven (Darani & Kaedi, 2017, Jing *et al.*, 2022, Kielarova & Sansri, 2016), and data-driven (Han et al., 2021) according to the computational mechanisms applied during the process. Experience-driven computation design usually applies reasoning methods such as case-based reasoning, ontology-based reasoning, productions rule-based reasoning, etc. to reuse explicit design knowledge and experiences. Model-driven computation design is usually conducted through design calculation or simulation optimization on the basis of the geometric (Kielarova & Sansri, 2016) /function (Darani & Kaedi, 2017) /mechanism models (Jing *et al.*, 2022) of the design objects. Data-driven computation design usually applies machine learning algorithms to extract implicit design knowledge and experiences hidden in design process data (Wang *et al.*, 2022). The relations among the concepts above are illustrated in Fig. 1.

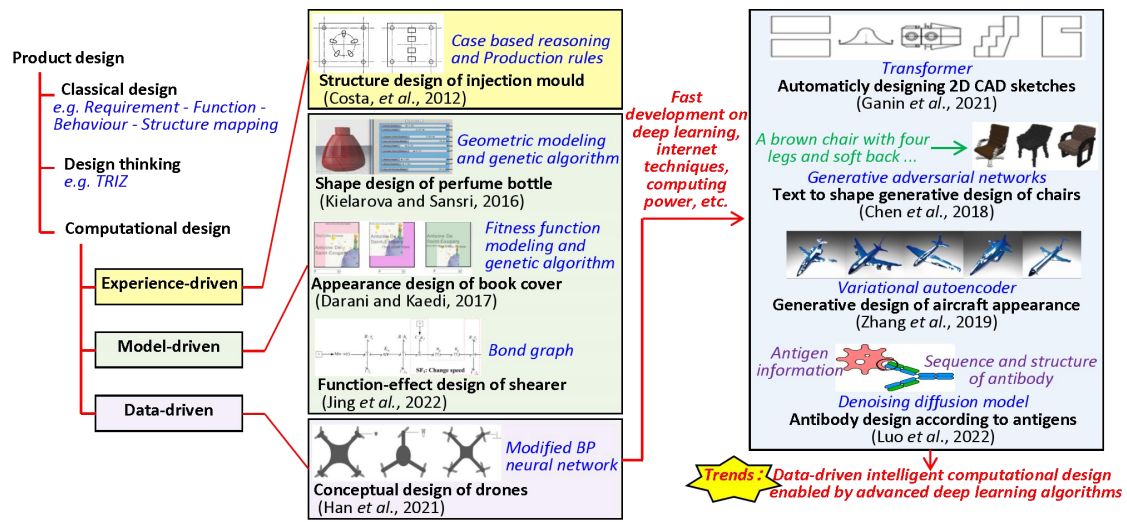

Fig. 1: Classification on product design methods.

Recently, artificial neural networks based multi-layer/multi-modal deep learning models have shown strong capabilities in complex data feature representation and pattern learning (Gan *et al.*, 2021, Yang *et al.*, 2022) and this provides the foundation for the application of artificial intelligence in data-driven product design from algorithm perspective. On the other hand, the fast development in large-scale computing servers, cloud-computing techniques (Buonamici *et al.*, 2020), high-speed internet based edge-cloud data collecting/storing/transmitting techniques (Ganin *et al.*, 2021), etc. provide the foundations from calculating power and training data perspectives. Under such background, utilizing the advanced deep learning algorithms for data-driven intelligent computation design (DICD) has become a research hotspot and has drawn the attentions from both academic and industrial fields (Krahe *et al.*, 2020). For example, *Google DeepMind* collected millions of CAD sketches from *Onshape* cloud CAD platform, and utilized the sketches as training data for a Transformer model which can support intelligent generation of CAD designs (Ganin *et al.*, 2021). Chen *et al.* (2018) developed a Generative adversarial networks (GAN) based generative design model which can generate 3D shapes according to text descriptions of required structural features. Zhang *et al.* (2019) developed a Variational auto-encoder (VAE) based generative design model for aircraft appearance generation. Gan *et al.* (2021) developed a generative design model for social robots based on an integration of Kansei engineering and GAN.

From the perspective of data science, product design requirements are usually expressed with text data which can be represented with relatively low-dimension data space, while design solutions are usually expressed with 2D/3D shape models containing design features such as measurements/accuracy/constraints/etc. and consequently they need to be represented with relatively high-dimension data space. In another word, the nature of design solution generation is cross-modal mapping from a point in the low-dimension design requirement data space with less information to a point in the high-dimension design solution data space with more information. Therefore, design solution generation process corresponds to a mapping path from the design requirement space to the design solution space, and a design solution corresponds to a point in the design solution space. **When non DICD methods are applied**, the mapping requires experiences from designers, white-box design models containing domain knowledge, simulation/real experiments, etc. to complete the evolution from a point with less information to a point with more information, and its complexity and time consumption are usually large. **When DICD methods are applied**, deep learning models can learn the patterns of the above mapping paths and the distribution of the points in the design solution space from large amount of historical design process data. Thus it enables fast and through search in the design solution space according to given design requirements. This makes DICD methods capable of effectively and intelligently generating large numbers of design solutions (Zhang *et al.*, 2019), lowing the threshold for executing the design process using the trained design models which have learned design knowledge from large numbers of historical design process data (Gan *et al.*, 2021), generating innovative design solutions that are difficult to acquire through non DICD design methods (Quan *et al.*, 2018), providing an automatic way to extract and reuse the implicit design experiences hidden in historical data (Zang *et al.*, 2022), etc.

However, as an emerging and frontier research topic, currently there are few systematic guidance on DICD implementation from full-process perspective. For example, 1) existing DICD researches usually focus on independent design problems within generative design and topology optimization (TO) steps, while the applications of DICD in the other steps during the entire product design process and the methods to realize full-process design intelligence are yet to be explored; 2) existing DICD researches usually paid more attentions to the algorithms yet relatively less attentions to the construction of the specific dataset which require feature engineering to narrow down the design features that are to be learned by the algorithms; 3) most of the existing researches seldom consider the explainability of DICD process from the viewpoint of conventional product design methods (e.g. the mapping relations among function, behaviour, structure, etc.).

In this regard, this paper aims at providing a systematic and operable road map for the development and application of DICD in the entire product design process, including a project planning workflow, an implementation framework, the soft computing mechanisms, key enabling technologies, and application scenarios of DICD.

## 2. Literature reviews

In this section, the literatures related to the implementation of DICD in different design steps have been reviewed.

### 2.1 Data-driven design solution retrieval

Compared with the commonly used methods for design solution retrieval, such as manual retrieval and simple reasoning algorithms based retrieval (e.g. production rule-based retrieval (Costa *et al.*, 2012)), data-driven retrieval has higher efficiency and accuracy when the historical design solution base is huge or the retrieval conditions are complicated (Zang *et al.*, 2022). The key for data-driven design solution retrieval is the feature representations of the retrieval conditions and the historical design solutions, and the similarity calculation between them (Li *et al.*, 2021).

According to the data structures of the retrieval conditions, data-driven design solution retrieval can be separated into single modal and multi-modal retrieval conditions (Patel *et al.*, 2021). **For single modal retrieval conditions**, Manda *et al.* (2021) developed a deep neural networks based model that can retrieve 3D CAD models according to freehand design sketches. Qi *et al.* (2021) developed a fine-grained retrieval model with attention mechanism, and the model is particularly effective when the design solutions are highly similar to each other in their structural features. Zang *et al.* (2022) developed an integrated retrieval model based on convolutional neural network (CNN) and recurrent neural network (RNN), and the model can identify the 3D shape models of mechanical parts according to text descriptions of required structural features. **For multi-modal retrieval conditions**, Vo *et al.* (2019) developed an image retrieval model that utilizes image and text data as retrieval conditions, and Ruan *et al.* (2022) developed a 3D shape retrieval model with joint representation of texts, 2D images, and 3D shapes.

## 2.2 Data-driven generative design

Generative design mainly refers to a kind of design method which utilizes prebuilt design rules, constraints, algorithm models, etc. to automatically generate the design solutions for give design tasks (Vlah *et al.*, 2020). According to the mechanisms that realize the generation of the design solutions, generative design can be roughly separated into four categories, including 1) shape grammar based approach (Alcaide-Marzal *et al.*, 2020), 2) parametric CAD modeling based approach (Khan & Awan, 2018), 3) evolutionary computation based approach (e.g. the genetic algorithm based generative design module by *Autodesk*) (Vlah *et al.*, 2020), and 4) data-driven approach.

The key for data-driven generative design is to utilize machine learning models to learn the patterns of the design solutions from historical design solutions, and then use the trained machine learning models to generate new design solutions for new design tasks (Regenwetter *et al.*, 2022). The commonly used machine learning models for data-driven generative design include GAN, VAE, Bayesian networks, etc. For example, Oh *et al.* (2019) developed a GAN based model that can generate 2D shapes of automobile hubs. Gan *et al.* (2021) developed a GAN based model for generating 2D conceptual images of social robots with different styles. Chen *et al.* (2019) developed a GAN based model for generating aircraft airfoils. Wu *et al.* (2022) developed a GAN based model for generating missile aerodynamic shapes. Nobari *et al.* (2021) developed a GAN based model for innovative bicycle design. Zhang *et al.* (2019) developed a VAE based model that can generate the appearance variations of air crafts. Kalogerakis *et al.* (2012) developed a Bayesian networks based model that can generate thousands of design solution variations from hundreds of historical design solutions. Luo *et al.* (2022) developed a denoising diffusion model based antibody design method which can generate the sequences and backbone structures of antibodies according to given antigen structures.

## 2.3 Data-driven topology optimization

Topology optimization is a kind of design optimization that aims at determining the most optimal distribution of the materials in a design space given predefined boundary and load conditions. Conventional TO methods are usually time consuming due to many rounds of iterative design calculation and large numbers of optimization variables during the calculation. In this regard many researchers are trying to use data-driven methods to accelerate the process of TO (Shin *et al.*, 2022).

Data-driven TO can be mainly separated into two types. **The first type** is to directly utilize data-driven models to acquire the TO results for the input conditions. For example, Abueidda *et al.* (2020) developed a CNN based model that can directly predict the TO results according to input constraints on boundary and load. Generative models such as GAN and VAE are also commonly used to directly produce TO results from input conditions, and these algorithms are capable of producing large numbers of alternative TO results with the similar patterns of the training samples (Nie *et al.*, 2021). Li *et al.* (2019) developed a two-stage pipeline in which the first stage utilizes a GAN model to predict low resolution near-optimal structure and the second stage utilizes another GAN model to acquire the refined TO results with high resolution. **The second type** is to combine data-driven methods with conventional TO methods, and it can be further separated into two sub-types. The first sub-type is to apply data-driven methods as part of the entire TO iteration process. For example, Hertlein *et al.* (2021) firstly utilized a GAN model to predict near-optimal TO results for additive manufacturing, and then applied a few rounds of conventional TO iterations on the basis of the results from the GAN model to filter the overhand structures which are not suitable for additive manufacturing. The second sub-type is to replace the time-consuming finite element analysis based objective function calculation and sensitivity analysis in conventional TO method with data-driven approaches (Xu *et al.*, 2021).

**2.4 Data-driven design evaluation**

Currently, the researches on data-driven design evaluation mainly utilize machine learning algorithms to separate the design solutions to be evaluated into predefined categories, each of which represents an evaluation grade. In most cases, 2D image is used as the data structure of the design solutions to be evaluated. For example, Tuinhof *et al.* (2019) developed a two-stage deep learning framework for garment design evaluation which can rank the fashion level of the garments according to their photos. Avola *et al.* (2022) developed an anomaly detection framework with CNN and convolutional auto-encoder, and it can identify defective products from good ones according to the 2D images of the products. Except for 2D images, Han *et al.* (2021) used a modified BP neural network to evaluate the conceptual design solutions of drones which are expressed in the form of design parameters. Li *et al.* (2021) developed a text classification based method for product evaluation according to the text comments from customers.

**2.5 Research Gaps**

It can be seen from above that DICD has recently become a research hotspot, and many researches have been devoted to utilizing data-driven methods to solve the problems during product design. However, there are still many issues to be considered for the further theorical development and industrial application of DICD. **Firstly**, as far as we know there lacks a systematic road map that introduces the methods and techniques to utilize data-driven models to improve the overall intelligence of product design process. **Secondly**, most of the researches paid more attentions about using certain kinds of algorithms to solve certain types of design related problems, yet seldom has clarified how to prepare the specific dataset and how to conduct design knowledge related feature engineering to identify the key design features that are supposed to be learned by the algorithm models. **Thirdly**, most of the researches on design solution generation are based on generative models which directly generate a group of design solutions from initial inputs. However, real product design process is usually conducted through many iterations each of which would make some updating on the previous iterations to gradually acquire a satisfied design solutions. Therefore how to apply data-driven methods for design solution updating according to gradually clarified design requirements from designers still requires further exploration. **Fourthly**, product design activities are highly related to Requirement - Function - Behaviour - Structure mapping, yet most of the researches skipped the analysis of this mapping relation and thus make the DICD researches unexplainable from the view point of conventional product design methods. **Fifthly**, there are still many issues in the entire product design process that could be supported with DICD methods yet have not been explored.

## 3. The general framework for DICD implementation

In this section, a general framework for DICD implementation is established, including a general workflow for DICD project planning, an overall framework for DICD project implementation, and the soft computing mechanisms involved in DICD implementation.

### 3.1 General workflow for DICD project planning

Before the detailed implementation of a DICD project, the designer should make a brief planning about the project by defining the input/output of the DICD model and select the suitable deep learning algorithms accordingly. The planning can be conducted through three main steps, as shown in Fig. 2.

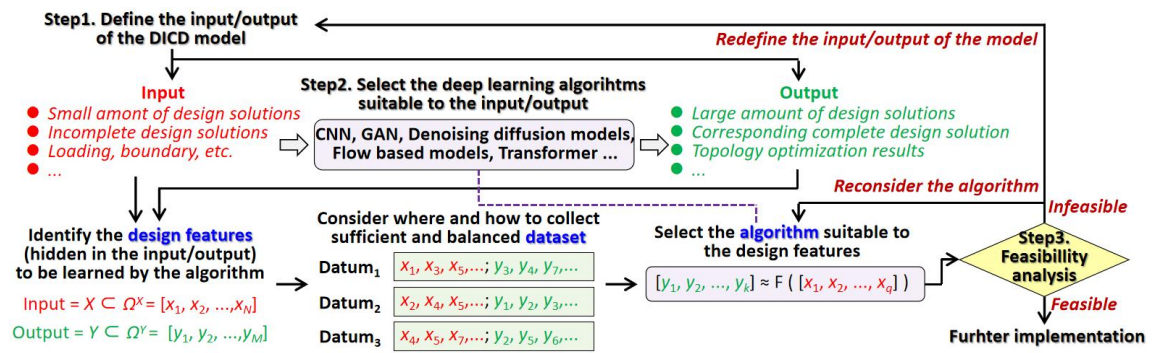

Fig. 2: General workflow for DICD project planning.

**The first step** is to clearly define the input and output of the DICD model, and a few examples are shown in Fig. 2. **The second step** is to select the deep learning algorithms suitable to the input and the output. The step can be achieved through three sub-step. Firstly, identifying the design features, hidden in the input and the output, that are supposed to be represented and learned by the deep learning algorithm. For example, if the DICD project is to generate 3D shapes of a mechanical part according to text descriptions, then the design features in the inputs (i.e. the text descriptions) would be the combinations of words and phrases that describe the functions or structures of the required 3D shapes, and the design features in the outputs would be the corresponding structural features in the form of 3D shape data (e.g. 3D voxel, 3D point cloud). Secondly, the designer need to consider how to collect sufficient and balanced data that contain the design features, such as collecting from historical design solution database, crowdsourcing, or data enhancing. Thirdly, selecting the suitable types of algorithms according to the types of the tasks (e.g. classification or generation) and the formats of the data (e.g. text or image). **The third step** is to evaluate whether enough data that contain the design features can be collected and whether the selected algorithm is suitable to the design features and the possible amount of data. Based on the evaluation, the designer may reconsider the entry point of the DICD project by redefining the inputs and outputs, or consider another type of algorithm.

### 3.2 Overall framework for DICD project implementation

After making a brief planning about the target DICD project by defining the input/output of the DICD model and select the suitable deep learning algorithms, the designer can implement the DICD project according to the overall framework illustrated in Fig. 3.

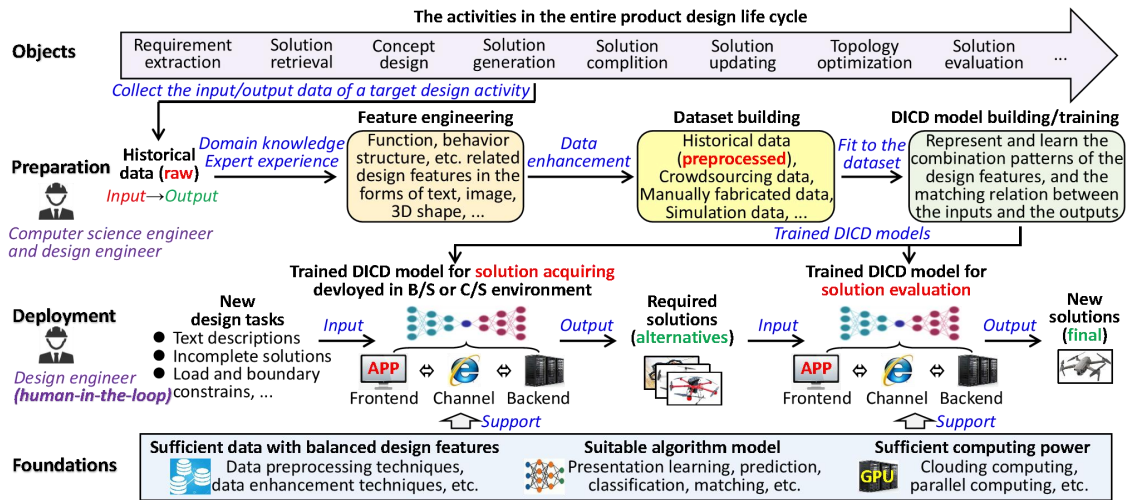

Fig. 3: The overall framework for DICD project implementation.

The entire framework can be described with four layers. **The first layer** contains all the activities during the entire product design process that can be supported with DICD models. After identifying a target design activity and its inputs and outputs, the designer can go to the second layer to prepare for the deployment of the DICD model. In **the second layer**, firstly the designer need to collect enough amount of historical data related to the target design activity. Then, feature engineering should be conducted to identify the design features that are supposed to be represented and learned by the deep leaning models from the perspectives of not only data structure but also function, behaviour, structure, etc. of the design solutions. Then the designer need to build a large enough and balanced dataset with preprocessed data for the later training procedure. Then the deep learning model suitable to the dataset should be fabricated and trained to achieve an acceptable accuracy for industrial applications. The nature of the training procedure is to let the fabricated deep learning model successfully represent and learn the combination patterns of the design features, and the matching relations between the inputs and the outputs. **The third layer** is for deploying the trained deep learning models to support DICD activities. The designer can deploy one or multiple trained models with different objects in the entire product design process. Each of the models should be conducted in human-in-the-loop manner, i.e. designer should participate the operation of the model by providing and adjusting inputs, making modifications on the intelligently generated solutions, making decisions on the basis of the generated solutions, etc. **The fourth layer** emphasizes on the foundations that support the deployment of the DICD models, including the dataset with sufficient and balanced design features, algorithm model suitable to the dataset, and sufficient computing power.

### 3.3 Soft computing mechanisms during DICD implementation

The computing process during DICD implementation is usually based on soft computing, and the general soft computing mechanisms for DICD models is illustrated in Fig. 4.

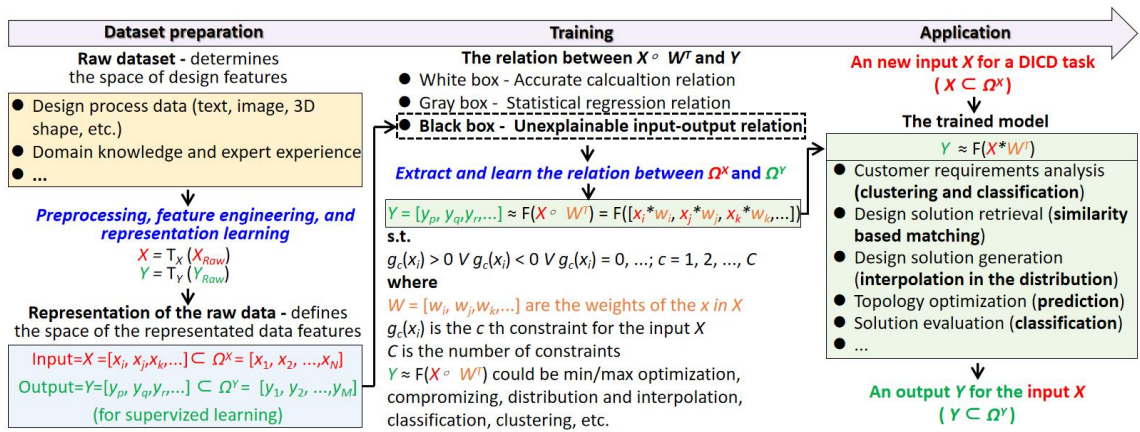

Fig. 4: The soft computing mechanisms during DICD.

During the procedure of **dataset preparation**, as shown in the left-side of Fig. 4, the key is to conduct representation learning on the raw dataset. The raw dataset, which contains the raw design process data, domain knowledge and expert experience, etc., defines the space of the design features (e.g. the function features, structure features). After preprocessing, feature engineering, representation learning, etc., the raw data

(denoted as $X_{Raw}$ and $Y_{Raw}$) would be transformed into more structured and featured data (denoted as $X$ and $Y$) that are easier to be analyzed in the modeling training procedure. Here, $\Omega^X = [x_1, x_2, ..., x_N]$ represents the space of the data features represented from the raw design data; $x_i$ represents a data feature in $\Omega^X$; and for each $X$ (which is the input of a piece of represented training datum), $X \subset \Omega^X$; $T_X$ indicates the overall function of preprocessing, feature engineering, and representation learning, etc. Similar definitions for $Y$, $y_j$, $\Omega^Y$ and $T_Y$ when the algorithm model is based on supervised learning. During the procedure of DICD **model training**, the key is for the DICD model to learn the relations between $\Omega^X$ and $\Omega^Y$. Generally, the deep learning based DICD models are black box models, therefore the relations can be expressed as $Y \approx F(X \circ W^T)$, where $W$ contains the weights of the data features in $X$, as shown in the middle of Fig. 4. The functions that are usually involved in $Y \approx F(X \circ W^T)$ includes min/max optimization, compromising solution, data distribution and interpolation, etc. During the procedure of DICD **model application**, the key is to utilize the trained DICD model to acquire the possible design solutions for the given input. A few examples of DICD tasks and their corresponding algorithm models are shown in the right-side of Fig. 4. It is worth mention that when the input $X \nsubseteq \Omega^X$, the trained DICD model may not be able to generate decent design solutions due to generalization problem.

**4. Key enabling technologies for detailed DICD implementation**

On the basis of the overall framework for DICD implementation, this section introduces the techniques that support the detailed implementation of DICD.

**4.1 Dataset building and feature engineering**

A decent dataset is the precondition for successful DICD project implementation. It defines the space of the design features that would be represented and learned by the algorithm model, and therefore it determines the upper limit of the performance of the DICD model from a sense (Halevy *et al.*, 2009). When building a dataset, design knowledge based feature engineering can be conducted to manually extract a group of explicit design features, and thus makes it easier for the algorithm model to represent and learn the patterns of the design features from the data. The technique for dataset building and feature engineering for DICD can be separated into seven steps, as illustrated in Fig. 5.

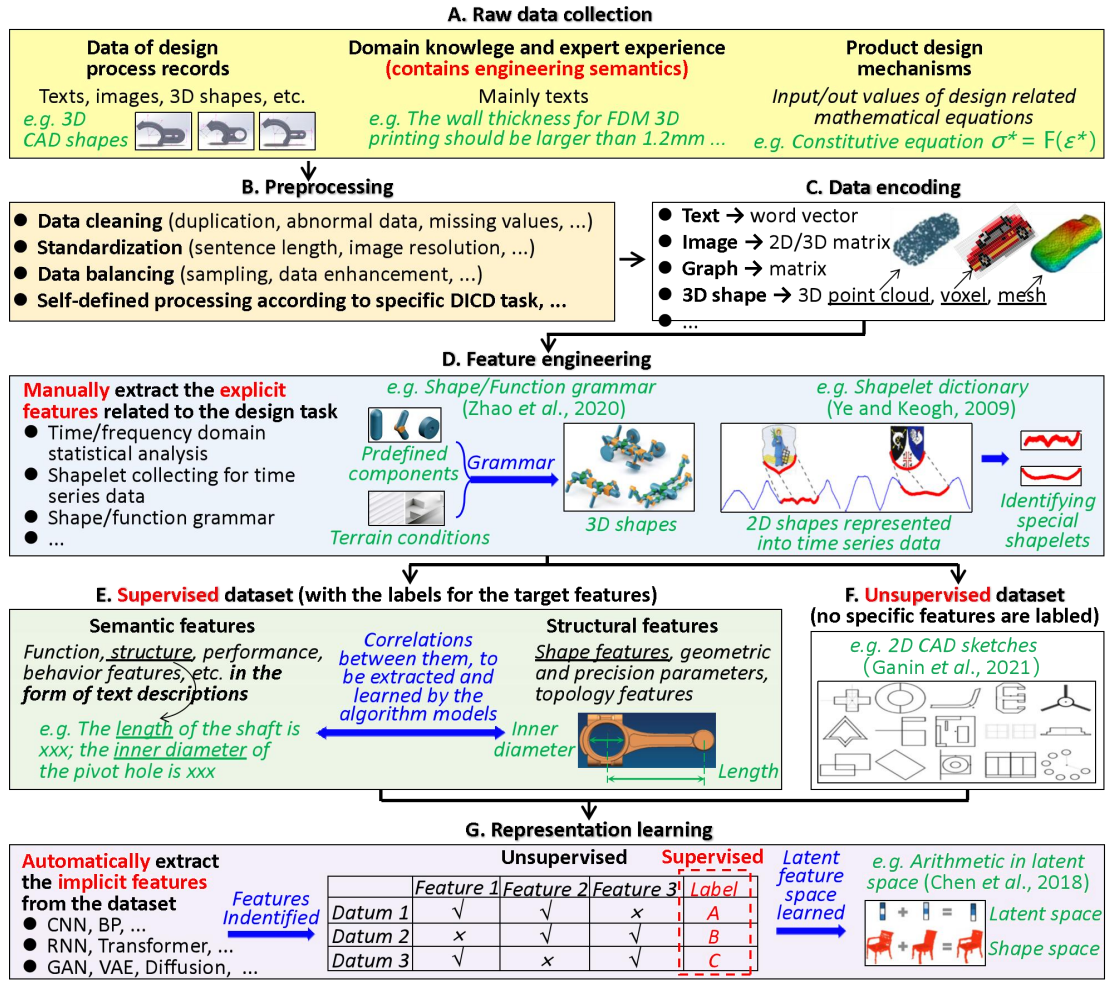

Fig. 5: The technique for DICD dataset building and feature engineering.

**A. Raw data collection.** Firstly, enough number of raw data should be collected, the data could be records of design process (e.g. design requirement descriptions, design drafts, 3D CAD models), domain knowledge and expert experience (usually expressed in text data and contain engineering semantics), and product design related mechanisms (in the forms of input/out of product design related equations and functions).

**B. Preprocessing.** After collecting the raw data, preprocessing should be conducted, including data cleaning, standardization, feature balancing, and self-defined processing according to the specific DICD task. During the process, crowdsourcing and data enhancing could be applied to guarantee enough number of data and balanced design features.

**C. Data encoding.** Sometimes the data have to be encoded into the formats that are easier to be processed by the algorithm models. For example, texts can be encoded into word vectors, images can be encoded into 2D/3D matrixes, 3D shapes can be encoded into 3D point cloud/voxel/mesh, etc.

**D. Feature engineering.** After preprocessing and encoding, feature engineering could be conducted as an additional step if needed. Here, feature engineering indicates manually identifying the explicit design features contained in the data, and thus makes it easier for the algorithm models to learn the relations among the design features. For example, shape grammar can be used to represent the key design features in the 3D shapes of target products (Zhao *et al.*, 2020). Example 2, after transforming 2D images into sequential data, shapelet dictionary can be built in which each shapelet represents a key shape feature (Ye & Keogh, 2009). It is worth mention that feature engineering is especially important when the amount of data is relatively small, as it can narrow down the design feature spaces that are supposed to be learned by the algorithm models

**E/F. Supervised/unsupervised dataset building**. After the previous steps, the dataset for model training can be built. The main difference between the two types of dataset is whether to provide manually added labels for each piece of datum. For supervised dataset, the labels are usually added as text data that contain semantic features (an example of supervised datum for design solution retrieval is illustrated in Fig. 5-E).

**G. Representation learning.** Here, representation learning indicates utilizing algorithm models to automatically extract the implicit features of the data. Commonly used algorithms include CNN (which is good at representing image data), Transformer and RNN (which are good at representing sequential data), etc. After representation learning, the features in each piece of datum would be identified, the combination pattern of the features learned, the feature space identified, based on which further DICD related computing can be conducted.

## 4.2 Data-driven design solution retrieval

The inputs of the retrieval model are retrieval conditions, which can be expressed with single modal data or multi-modal data, as shown in Fig. 6. The data formats of single modal retrieval condition includes text, 2D image, free-hand 2D/3D sketch, etc. Among them, text-based (Zang *et al.*, 2022) and sketch-based (Qin *et al.*, 2022) retrieval can be particularly supportive for industrial applications, and it would improve the retrieval efficiency and accuracy if the text descriptions are provided according to feature engineering which defines the key design features of the target types of design solutions (Zang *et al.*, 2022). Multi-modal retrieval indicates using different types of data format as retrieval conditions at the same time. For example, the designer can draw a free-hand sketch and provide a few sentences as supplement description of the retrieval condition to improve the retrieving accuracy.

The outputs of the retrieval model are usually the original 2D/3D CAD models. However, the calculation cost of directly predicting the similarity between retrieval conditions and the CAD models are high, therefore the original CAD models (especially the 3D CAD models) are usually compressed and encoded into lightweight data formats during the retrieving (e.g. 3D voxel, point cloud).

During the retrieving, firstly the retrieval conditions and historical solutions would be transformed into feature vectors in relatively lower dimension space compared to the original input and output data, and then both supervised similarity prediction models and unsupervised clustering models can be utilized for retrieval. For supervised learning based retrieval, the input and the output would be transformed into spaces with different dimensions. For unsupervised leaning based retrieval, the input and the output would be transformed into the same space. Secondly, the similarity of the retrieval condition and the historical solution would be predicted through fitting functions in supervised leaning models and through clustering in unsupervised learning models.

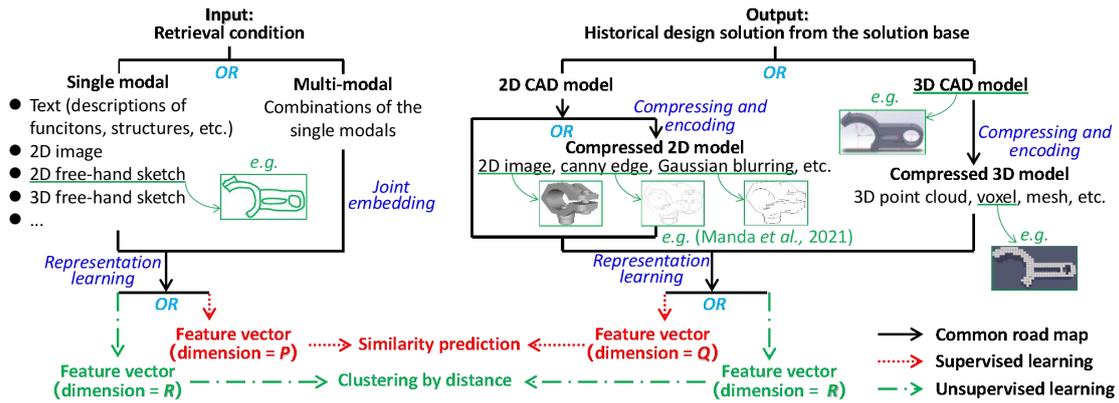

Fig. 6: The technique for data-driven design solution retrieval.

**4.3 Data-driven design solution generation, transformation, and completion**

The core idea for data-driven design solution generation, transformation, completion is to firstly use algorithm models to extract and learn the combination patterns of the design features from historical design solutions (i.e. the training stage), and then use the trained algorithm models to generate new design solutions for new design tasks. In this subsection, three typical techniques for data-driven design solution generation/transformation/completion are introduced, as illustrated in Fig. 7.

The first technique is for generating new design solutions according to text descriptions using conditional generative models such as conditional GAN. The solutions could be conceptual design drafts of 2D image or 3D shape, and the text descriptions may describe the required solution from the perspectives of function, structure, etc. An example is illustrated in Fig. 7-A. During the training stage, each piece of text description would be jointly embedded with random noise (denoted as $z$) and then transformed through a Generator into feature vectors, and the feature vectors represent the text description and a corresponding fake design solution. Each pair of real solution and its corresponding text description from dataset would be jointly embedded through an Encoder into feature vectors, and the feature vectors represent the text description and its corresponding real design solution. The Discriminator would be trained to identify the two types of feature vectors until it cannot tell the differences. After been trained, the Generator would be able to generate new design solutions according to input text descriptions for new design tasks. It is worth mention that the training strategy here is only a simple example. More complicated and advanced training strategies can be established to achieve better generation performance (Reed *et al.*, 2016).

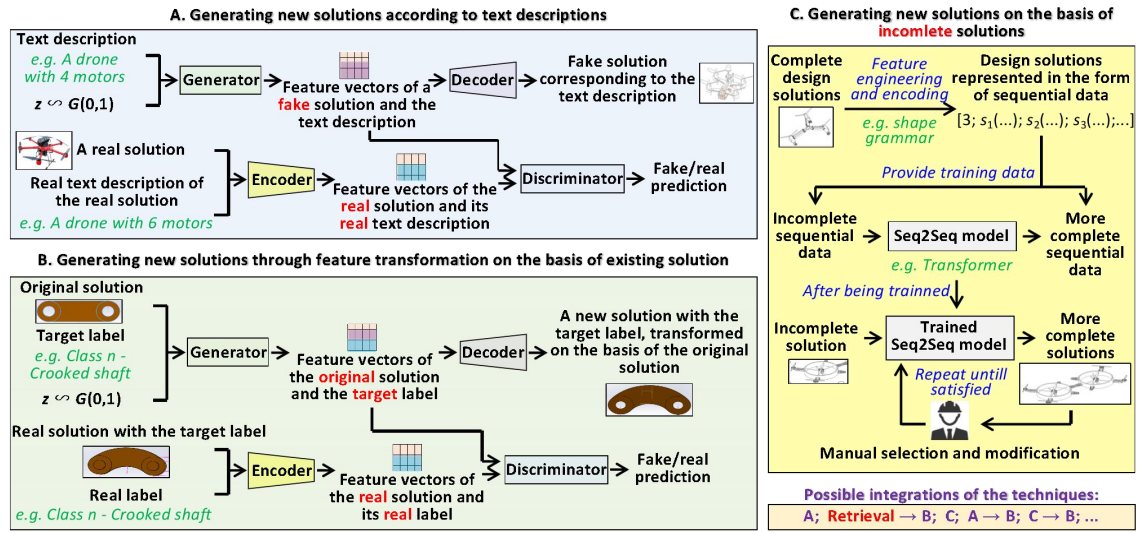

Fig. 7: Typical techniques for data-driven design solution generation, transformation, and completion.

The second technique is for generating new design solutions through feature transformation on the basis of existing solutions, and it can be also achieved using conditional generative models. An example is illustrated in Fig. 7-B. The example task is to generate a connecting rod model with crooked shaft on the basis of an existing connecting rod model with straight shaft. During the training stage, original solution (e.g. a straight-shaft rod), label that represents the class of the solutions to be transformed to (e.g. crooked shaft), and random noise *z* would be jointly embedded into feature vectors through a Generator. Solution of the class that the original solution is supposed to be transformed to (e.g. a crooked-shaft rod) and its class label (e.g. crooked shaft) would be jointly embedded into feature vectors through an Encoder. The Discriminator would be trained to identify the two types of feature vectors until it cannot tell the differences. After been trained, the Generator would be able to transform an existing solution to a target type of solution defined by the input label.

The third technique is for generating new design solutions by intelligently providing suggestions on the basis of existing incomplete solutions. The technique can be achieved through three steps. Firstly, represent the design solutions in the form of sequential data. For example, an design solution in the form of 2D image or 3D shape can be represented with a list of parameterized grammar rules using shape grammar (Zhao *et al.*, 2020). Secondly, using Seq2Seq models (e.g. RNN, Transformer) to learn the combination and sequential relations among the grammar rules using the sequential data from real design solutions as training data. After being trained, the Seq2Seq model would be able to provide suggestions about the next pieces of grammar rules that can be combined together with the input sequence data of grammar rules. Thirdly, using the trained Seq2Seq model to give suggestions about how to complete the input incomplete design solution, and the designers can select from the suggestions and make modifications if needed, and repeat the procedure until completed design solutions have been generated.

It is worth mention that the three techniques above can be applied independently or integratedly, as shown in the bottom right in Fig. 7.

## 4.4 Data-driven topology optimization

Data-driven TO techniques can be enabled with prediction models or generation models. Five typical data-driven TO techniques and their integration approaches are illustrated in Fig. 8.

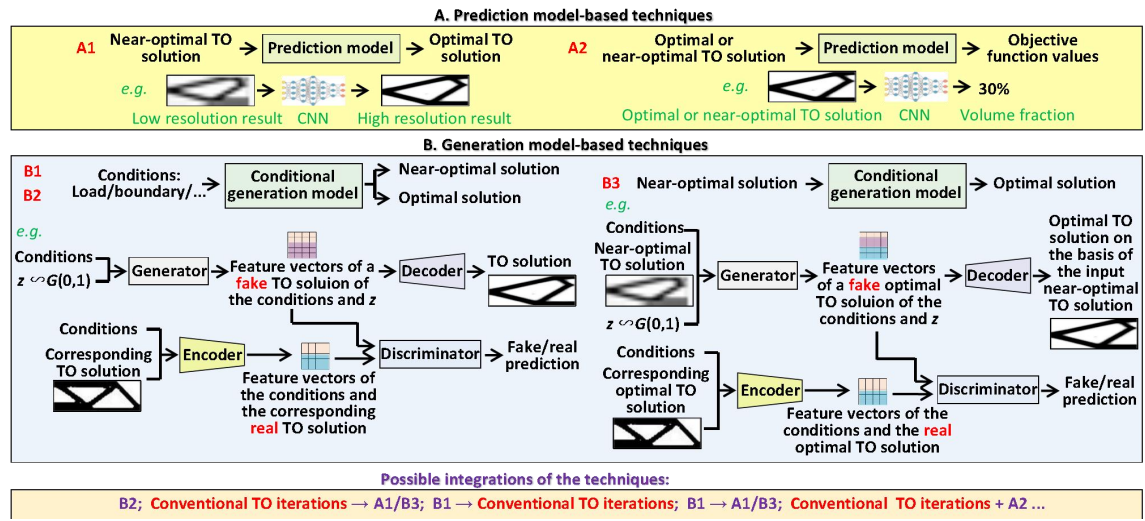

Fig. 8: Typical techniques for data-driven TO.

**A1. Predicting optimal TO solution according to near-optimal TO solution.** Here, near-optimal TO solutions can be acquired through a few rounds of conventional TO iterations, and by integrating data-driven TO with conventional TO, the total time cost for TO can be reduced. An example of A1 is illustrated in Fig. 8-A1, in which a supervised CNN based model is utilized to represent and learn the mapping relations between low resolution TO results and high resolution TO result.

**A2. Predicting objective function values according to the input of optimal or near-optimal TO solution.** For example, supervised CNN can be trained to predict the compliance values or volume fraction values for input TO solutions.

**B1/B2. Generating near optimal/optimal TO solution according to TO conditions.** The core idea for these two techniques is to utilize conditional generative models (e.g. conditional GAN/VAE) to learn the combination patterns of the 2D/3D shape features in the TO solutions, together with the mapping relations between the patterns and the TO conditions (e.g. load, boundary). An example is illustrated in Fig. 8-B1/B2. During the training stage, the feature vectors that represent the conditions and random noise $z$ would be generated by a Generator; the feature vectors that represent the conditions and their corresponding TO solution would be jointly embedded by an Encoder; the Generator would be trained until the Discriminator cannot tell the difference between the two types of feature vectors. After been trained, the Generator would be able to generate TO solutions given input TO conditions.

**B3. Generating optimal TO solution according to near-optimal TO solution.** B3 can be applied to acquire TO solutions with higher resolution, clearer and smoother boundaries, without undesired structural features, etc. on the basis of input near-optimal TO solution. Its training strategy is similar to B1/B2, however the contents of the training data are different, as illustrated in Fig. 8-B3.

The five data-driven TO techniques can be integrated with conventional TO techniques. A few possible integration approaches are listed in the bottom of Fig. 8.

## 4.5 Data-driven design solution evaluation

Data-driven design solution evaluation can be enabled with one-dimension classifier, multi-dimension classifier, and anomaly detection model, as shown in Fig. 9.

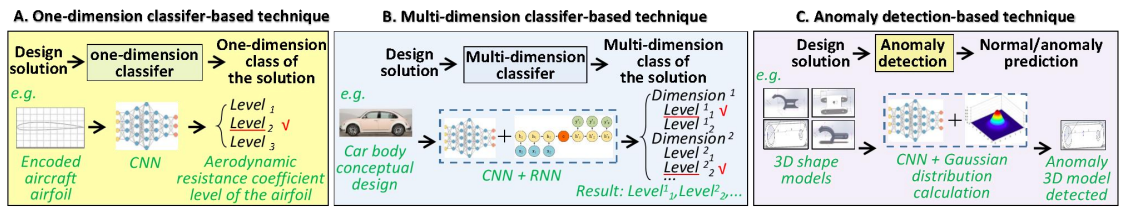

Fig. 9: Typical techniques for data-driven design solution evaluation.

**A. One-dimension classifier-based technique.** The technique can evaluate a design solution from one dimension, and the design solution to be evaluated can be 2D image, 3D shape model, graph model, design parameter combination, etc. By applying suitable one-dimension classifier algorithm, the design solutions can be separated into different classes, each of which represents a predefined evaluation result. For example, aircraft airfoil design solutions encoded in the form of 2D image can be separated into multiple classes with CNN-based models, and each class may represent a range of aerodynamic resistance coefficient of airfoils.

**B. Multi-dimension classifier-based technique.** In some scenarios, design solutions need to be evaluated from multiple dimensions, and in this kind of scenario Seq2Seq models can be applied. For example, conceptual design solutions of car bodies encoded in the from of 2D image or 3D shape can be separated into multi-dimension classes with integrated model of CNN and RNN. The evaluation results are represented in the from of sequential datum, in which each value in the sequential datum represents an evaluation level from a dimension, as illustrated in Fig. 9-B.

**C. Anomaly detection-based technique.** The two techniques above are for the scenarios when the classes of the evaluation results can be clearly defined, and when balanced labeled training data can be collected for each class. However, in some scenarios the evaluation task is only to separated the design solutions into normal and anomaly ones, and there are too many classes of anomaly solutions, and it is difficult to collect balanced labeled training data for the anomaly solutions. In this kind of scenario, anomaly detection-based evaluation technique can be applied. For example, when the task is to evaluate the automatically generated design solutions from generative models, normal solutions can be used to train the anomaly detection-based evaluation model. After been trained, the evaluation model can identify any types of of anomaly design solutions from large number of input solutions.

## 4.6 Dynamic knowledge graph based explainable DICD process and result representation

The techniques above are mainly focused on independent steps in the entire product design process, and the processing between the inputs and outputs of each of these steps are realized with deep learning based black-box models, and thus makes the processing unexplainable from the perspective of classical product design methods. Here a meta knowledge graph (KG), graph2graph generation, and graph2shape generation based approach is proposed to improve the overall explainability of DICD process, as illustrated in Fig. 10. An example of connecting rod design is used for demonstration. Firstly, a meta KG of connecting rod design is established (Fig. 10-a). Then, raw design requirements can be input by designers (Fig. 10-b), based on which intelligent entity/relation extraction can be conducted to extract the functional design requirements in the form of triplet data (Fig. 10-c). After that, graph2graph generation can be conducted to acquire the structural features of the connecting rod in the form of triplet data, and designers can make modifications on the triplet data to give more accurate definition of the connecting rod. Eventually, graph2shape generation can be applied to generate the 3D shape models of the connecting rod according to the KG in Fig. 10-e. In this way, both the design solutions and the mapping design process can be recorded with human/computer readable KG, and thus KG driven retrieval/matching/reasoning/timeline/etc. techniques can be applied to enable more intelligent tasks for product design supporting. The example here only applied the DICD techniques from Section 4.3, and more techniques can be integrated in the KG based approach for more complicated DICD project.

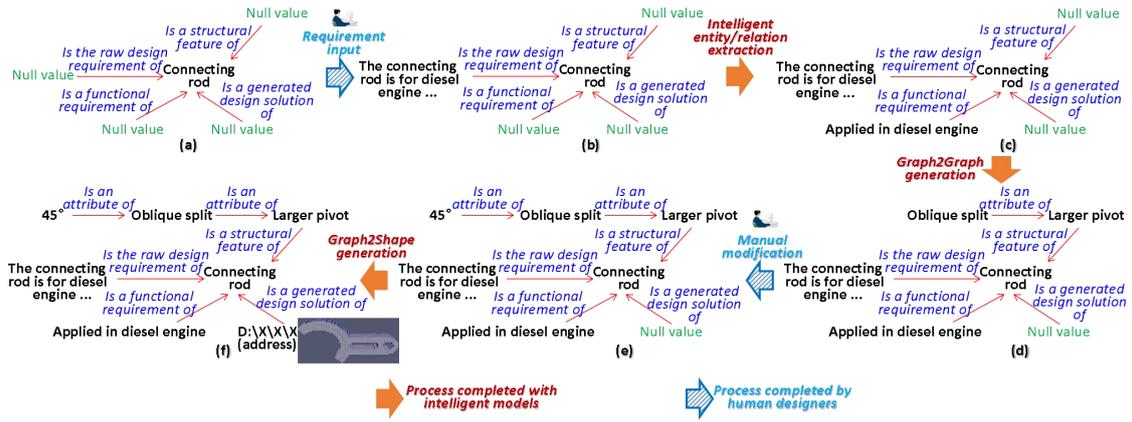

Fig. 10: Dynamic knowledge graph based connecting rod design process and result representation.

## 5. Application scenarios and case studies

On the basis of the universal DICD methods and techniques in Section 3 and 4, three example application scenarios of the methods and techniques are introduced below.

### 5.1 Linking rod design solution retrieval and generation

Linking rod is an important mechanical part widely applied in link mechanism. Here, a case study of data-driven linking rod retrieval and generation models is used to demonstrate the application of DICD in mechanical part design, as illustrated in Fig. 11.

**Firstly**, feature engineering is applied to determine a group of key structural features and feature values of linking rods. For example, *Shaft* is a component of linking rod, *Cross section* is a structural feature of *Shaft*, *H-shape* is a feature value of *Cross section* (Fig. 11-A). **Secondly**, the 3D CAD models of 15 typical types of linking rods were drawn with *Solidworks*. Each of the CAD models were modified into around 80 variants by permutation and combination the feature values of the structural features, and totally more than 1000 variant 3D CAD models of the 15 typical types of linking rods were fabricated (Zang *et al.*, 2022). Each of the variants were attached with a paragraph of text descriptions that describe the structural features of the variant (Fig. 11-B). **Thirdly**, an integration model of CNN-RNN was used for text representation learning and a 3D CNN model was used for 3D shape representation learning. On this basis, text2shape retrieval model and generation model were built with cross modal similarity calculation mechanism and conditional GAN, respectively. Here, the original 3D CAD models were transformed into 3D voxel models to train the text2shape retrieval/generation models. **Fourthly**, after been trained the text2shape retrieval/generation models can be applied in the scenario of case-based reasoning driven linking rod design. After inputting a paragraph of texts describing the structural features of a new design task, the retrieval model can retrieve the 3D CAD models most suitable to the new design task from historical database, and if satisfying solutions can not be retrieved, which means there are not similar historical design solutions in the database, then the generation model can be used to generate new 3D shapes to provide reference for the design task.

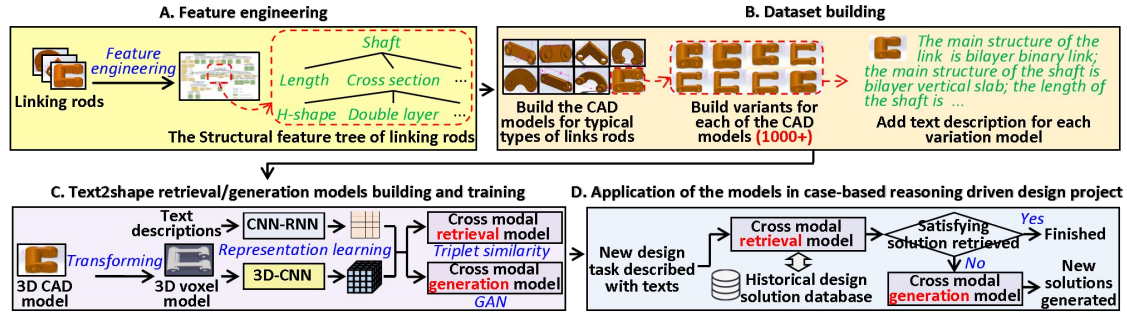

Fig. 11: Application of text2shape retrieval/generation in case-based reasoning driven linking rod design.

Compared with the other retrieval/generation model applications in product design, the case here emphasizes on systematic structural feature analysis of the design target during representation learning (Zang *et al.*, 2022), and thus it can improve the retrieval/generation accuracy from the perspective of product design function - structure mapping. The text descriptions in this case were organized in the form that can be easily transformed into triplet data. For example, "*the main structure of the shaft is bi-layer vertical slab*" can be transformed into <*Bi-layer vertical slab*, *is the main structure of*, *Shaft*>. The purpose is discussed in Section 6.3.

**5.2 Aircraft wing shape design solution generation and evaluation**

Conventional approaches for aircraft wing shape design require building complicated white-box design models, time-consuming optimization calculation and computational fluid dynamics evaluation experiments, and it usually repeat many iterations before acquiring satisfying solutions (Chen, 2019).

In this regard, a DICD approach driven with conditional text2shape generation model and vector sequence prediction model is established, as shown in Fig. 12. The entire process can be separated into three stages. **Firstly**, historical design solutions and their function and aerodynamics performance information can be used to fabricate the training data for the two models (marked with dot lines in Fig. 12). During the process, feature engineering and prepossessing are required to change the original historical design solutions into the formats that are easier for the deep learning models to analyze. **Secondly**, the conditional text2shape generation model and the vector sequence prediction model are applied for generating aircraft shape models according to input design requirements, and then evaluating the aerodynamic characteristics of the generated aircraft shapes (marked with solid lines in the figure). Here, denoising diffusion model are suggested for the generation model due to its capability of generating aircraft shapes with fine-grained structure features, and Transformer model is suggested for the vector sequence prediction model due to its capability of analyzing and learning the mapping relations between long/complicated/interrelated vectors of aerodynamic evaluation criteria. **Thirdly**, after acquiring satisfying design solutions from the deep learning models, final design and test stage can be applied, and the final solutions can be used to extend the training dataset (marked with dot dash lines).

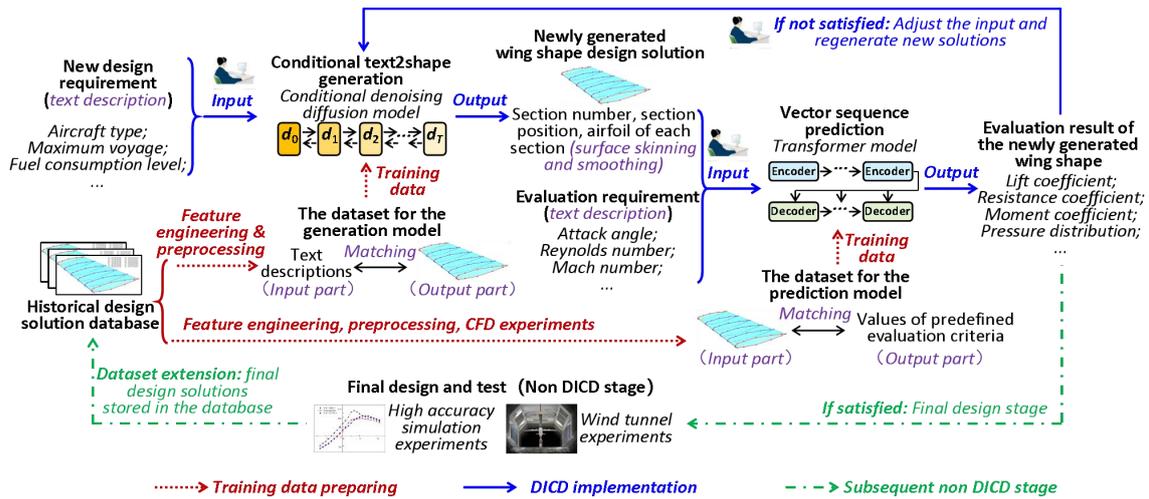

Fig. 12: Application of text2shape generation and vector sequence prediction models for aircraft wing shape design.

Compared with other data driven methods for aircraft wing shape design, the denoising diffusion models and Transformer models applied in this case can learn the fine-grained mapping relations among the structure features of wing shapes and the functional/performance features hidden in the engineering semantics.

**5.3 Car body interactive conceptual design**

The conventional approach for car body design can be roughly separated into five steps, which are 1) drawing the 2D drafts, 2) building clay model, 3) conducting wind tunnel experiments and modifying the clay model accordingly, 4) 3D scanning the clay model to acquire 3D CAD model, 5) optimization design on the 3D CAD model to acquire the final result. Here, a text2shape representation learning and mapping based DICD method is established, it can be applied to support innovative design during Step 1, and it may replace Steps 1-4 for low-price electric cars which has larger innovative potential for body design yet do not require strict wind tunnel experiments as they are usually low speed and light load.

The key idea of the DICD method here is to represent the latent space of the design features in the design requirement engineering semantics and the shape space of car body, and learn the mapping relations between the two. Conditional generative models can be used to realize the DICD model, and here we use condition VAE for demonstration, as illustrated in Fig. 13. **During the training stage**, firstly historical 3D CAD models of car bodies would be changed into 3D surface point cloud models, and each of the models would be added with a label which indicates its scores in each predefined design features. For example, in the bottom left of Fig. 13, the design features are defined according to Kansei engineering dictionary. Secondly, conditional VAE model can be trained using the labeled point cloud models. After been trained, the decoder of the conditional VAE would be able to generate point cloud models according to input conditions (i.e. the labels). **During the application stage**, the trained decoder could be encapsulated into interactive design APP. After the designer had input the design requirements by determining the scores for each design features (i.e. sampling in the latent space), the trained decoder would be able to generate corresponding 3D point cloud models which can be quickly rendered into 3D mesh, voxel, etc. to visualize the design results. Then the designer can making adjustments by changing the input scores until a final result had been generated.

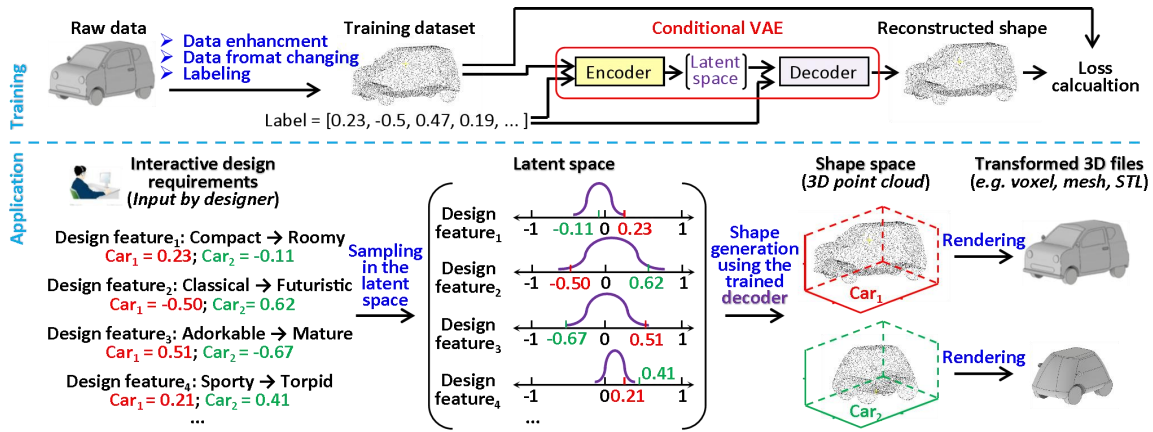

Fig. 13: Application of text2shape representation learning and mapping for car body interactive conceptual design.

Compared with the other generation model applications in product design, the case here stands out by enabling interactive design between designer and intelligent design model. After adjusting the input design requirements through the dragging bars, each of which is related with one of the design features, the intelligent design model can quickly respond by generating a corresponding design solution, and then the designer can check the generated solution and readjust the input design requirements if necessary.

## 6. Discussion and conclusion

### 6.1 Contribution

The main contribution of the paper is a systematic and operable road map for DICD implementation from full-process perspective, including a general workflow for DICD project planning, an overall framework for DICD project implementation, the soft computing mechanisms during DICD, six key enabling technologies that support the detailed implementation of DICD, and three examples of DICD application scenarios. Compared with the existing researches on DICD, the road map considers not only the intelligence for independent design steps but also the intelligence for the entire product design process; it pays attentions to not only the applications of deep learning models from computer science perspective, but also specific dataset building and design knowledge related feature engineering to narrow down the design feature spaces that are to be learned by deep learning models; it explores the way to improve the overall explainability of DICD process from the viewpoint of conventional product design methods by visually representing the mapping relations and results among raw design requirements, functional requirements, structural features, etc. Supported with the road map, the common mechanisms and calculation principles of existing DICD researches can be revealed, based on which the possible DICD applications that have not been explored can be systematically guided.

**6.2 The issues to be further explored**

6.2.1 How to support the industrial applications of DICD

In a nutshell, DICD is to use advanced deep learning algorithms to extract and represent and design features hidden in historical or fabricated design process data, and then learn the combination and mapping patterns of these design features for the purposes of design solution retrieval, generation, optimization, evaluation, etc. Like any other deep learning application projects, the success of a DICD project depends on rational coordination among dataset, algorithm, and computing power. However, DICD is relatively less developed compared with deep learning driven intelligent production, operation, service, etc. We believe that the main reason is because it is more difficult to build the specific datasets for specific DICD tasks, as the production/service data can be efficiently collected with industrial internet of things, cyber-physical systems, etc. but the design process data can not. It is also worth mention that the researchers in computer science have built many useful datasets for text2shape, image2shape tasks, etc. (Chen *et al.*, 2018), and these datasets cover many commonly used data formats such as 2D image, 3D voxel/point cloud/mesh. However, these datasets cannot be directly supportive for DICD researches and applications, as they usually do not consider engineering semantics nor engineering design features. Therefore, the development and application of DICD should devote more to the building of the datasets containing the targeted engineering semantics and design features. Strategies such as crowdsourcing, data enhancement, etc. could be applied to collect more raw data; transfer learning, small dataset learning algorithms might be helpful for reducing the number of required training data; and feature engineering-based preprocessing should be applied to improve the efficiency for model training.

6.2.2 How to utilize the suitable algorithm model for DICD

As discussed before, the success of DICD requires the coordination between the dataset and the algorithm. In the past decade, large numbers of different types of deep learning algorithms and their variants have been developed. A DICD engineer need to select the most suitable algorithm for his design project from all those algorithms, and the key is to firstly transfer the target design problem into a deep leaning problem (e.g. image generation for TO, image classification for design evaluation), and then select the suitable algorithm according to the input/output of the design problem. There are some basic guidance for algorithm selection (e.g. CNN is suitable for image data processing, RNN/LSTM are suitable for text data processing), however, which specific variant of a type of algorithm should be used should be determined by the design features in the specific dataset for the specific DICD project. Therefore parallel comparison should be conducted rather than directly select an algorithm and its variant according to the guidance from computer science researches. On the other hand, some of the recently established algorithms from computer science have shown great application potential yet have not been fully explored for DICD. For example, denoising diffusion models (Luo *et al.*, 2022) and flow-based generative models (Xu *et al.*, 2022) have shown stronger capability for generating high resolution 2D/3D shapes and therefore may replace GAN/VAE in certain application scenarios, and reinforcement learning models based design solution generation and optimization could also be a promising direction. However, the algorithms from computer science may not be capable of directly application in engineering design tasks. For example, the existing conditional generation models are usually for small number of discrete conditions, while in engineering design tasks the design requirements usually contains large numbers of continuous data as conditions. Therefore further fabrications should be conducted on the algorithm models accordingly.

**6.3 Future works**

Our future works would be conducted from the following directions: 1) exploring more entry points in the entire product design process for the application of DICD; 2) developing the methods for building the specific dataset for engineering design tasks, together with the feature engineering-based data preprocessing methods that can improve the training efficiency and performance of the DICD deep learning models; 3) taking advantages of the most advanced deep learning algorithms from computer science researches, and establishing an operable guidance for algorithm selection and modification for DICD tasks; 4) further improving the explainability of DICD by integrating conventional design methods into DICD process.

**Credit author statement**

Maolin Yang: Method, Writing-Original Draft. Pingyu Jiang: Conceptualization, Method, Writing-Review & Editing, Supervision, Resources. Tianshuo Zang: Software. Yuhao Liu: Validation.

**Acknowledgement**

This work was supported by National Natural Science Foundation of China (No. 51975464); Natural Science Basic Research Program of Shaanxi Province (No. 2023-JC-QN-0397).